\documentclass[runningheads,a4paper]{llncs}

\usepackage{amssymb}
\usepackage{verbatim}
\usepackage{float}
\usepackage{placeins}
\usepackage[final]{graphicx}
\usepackage[export]{adjustbox}
\usepackage[caption=false,font=footnotesize]{subfig}
\usepackage[normalem]{ulem}
\usepackage{soul,xcolor}
\usepackage{multirow,tabularx,booktabs,ctable}
\useunder{\uline}{\ul}{}

\let\belowcaptionskip\abovecaptionskip
\usepackage{url}
\urldef{\mailsa}\path|{roi, acm, bek}@zurich.ibm.com|
\urldef{\mailsb}\path|{ristrate01, d.nikolopoulos}@qub.ac.uk|
\newcommand{\keywords}[1]{\par\addvspace\baselineskip
\noindent\keywordname\enspace\ignorespaces#1}

\begin{document}
\newcommand{\comments}[1]{\textcolor{blue}{[#1]}}
\newcommand{\change}[1]{\textcolor{red}{#1}}
\setstcolor{red}
\newcommand{\removetext}[1]{\st{#1}}
\newcommand{\rephrased}[1]{\textcolor{magenta}{#1}}
\newcommand{\added}[1]{\textcolor{violet}{#1}}
\newcommand{\gmadd}[1]{\textcolor{blue}{#1}}
\newcommand{\gmrm}[1]{\sout{#1}}

\mainmatter

\title{Incremental Training of\\ Deep Convolutional Neural Networks}

\author{R.\ Istrate\inst{1}\inst{2}
\and A.\ C.\ I.\ Malossi\inst{1}
\and C.\ Bekas\inst{1} 
\and D.\ Nikolopoulos\inst{2}}
\authorrunning{R. Istrate, A.C.I. Malossi, C. Bekas, D. Nikolopoulos}

\institute{IBM Research -- Zurich, Switzerland\\ \mailsa
\and Queen's University of Belfast, United Kingdom\\ \mailsb}

\maketitle

\begin{abstract}

We propose an incremental training method that partitions the original network into sub-networks, which are then gradually incorporated in the running network during the training process. To allow for a smooth dynamic growth of the network, we introduce a look-ahead initialization that outperforms the random initialization. We demonstrate that our incremental approach reaches the reference network baseline accuracy. Additionally, it allows to identify smaller partitions of the original state-of-the-art network, that deliver the same final accuracy, by using only a fraction of the global number of parameters. This allows for a potential speedup of the training time of several factors. We report training results on CIFAR-10 for ResNet and VGGNet.

\end{abstract}

\section{Introduction} \label{sec:introduction}
When dealing with a classification task on a new dataset of images, a widely used strategy is to start by training a few state-of-the-art networks that were developed for similar datasets. The main disadvantage of this approach is that only too late in the process we learn whether the network is not well suited for the dataset. At that moment, the training is stopped, the network is adapted and the process is restarted from scratch, discarding all the previously collected information. Thus, in the presence of large, complex datasets, the global time-to-solution tends to reach the order of several weeks or even months. 

A great deal of research is now focused in optimizing the depth of a neural network (NN) and most of the proposed methods involve growing the network by gradually adding one or more layers \cite{adanet,jung,discriminative}. Although this is a logical step towards evolving network architectures, to the best of our knowledge there is no work that compares the accuracy and performance of the networks obtained by dynamically adapting their structure during training, with the original one trained from scratch. 

In this paper we present our contribution towards gradually training deep state-of-the-art convolutional neural networks (CNNs) with no loss in accuracy.
During incremental training we start the learning process with a shallow network. When the network performance stops improving because of its limited capacity, we transfer the knowledge of the shallow network to a deeper one and continue the training. In this way the time and resources spent in training the initial shallow network are not wasted and the deeper network has a better initialization point. Additionally, since the incremental training method easily quantifies (by construction) the value added by each network extension, it is simple to define custom trade-off functions (e.g., time vs accuracy) to stop the network expansion when desired criteria are met. The long-term goal is to develop a fully-automated framework that optimizes neural network structures for specific tasks. However, this requires knowing what type of changes need to be performed to the network, which is outside the scope the current work.

The rest of the paper is organized as follows. Section~\ref{sec:relatedwork} briefly summaries related works, Section~\ref{sec:methodology} provides an overview of our methodology, while Section~\ref{sec:experiments} presents  the results of our experiments. Finally, concluding remarks and future works are summarized in Section~\ref{sec:conclusion}.

\section{Related work} \label{sec:relatedwork}

There is a clear tendency in the literature towards automatizing the design of NN. In AdaNet~\cite{adanet} the authors aim to optimize the network architecture and the internal weights, by balancing the trade-off between model complexity and empirical risk minimization. Although the claim is that networks discovered through this approach perform better than those found with a grid search, only limited results for binary classifiers trained on subsets of CIFAR-10~\cite{cifar10} dataset are presented.

Auto-Net \cite{autonet} is a framework that automatically tunes feed-forward neural networks without human intervention. The authors focus only on fully-connected neural networks in order to keep the number of hyper-parameters at a manageable level. They tune 63 network and layer dependant hyper-parameters for networks which are at most 6~layers deep. With this method they generated an ensemble of 39 models that outperformed all human experts and was the first automatically generated model to win an image competition~\footnote{http://automl.chalearn.org/}.

On the same track, a framework for large scale image classifiers~\cite{evolution} makes use of genetic intuitive mutations to explore unprecedented large search spaces. The fully automatic evolution culminates with a trained network that reaches 94.6\%~accuracy on CIFAR-10, 2\%~lower than manually engineered state-of-the-art networks. However, the prohibitive computational cost of the framework, that required $4\cdot 10^{20}$~FLOP for a 12~layer network, makes this approach unfeasible in practical scenarios with bigger datasets. 

Other works \cite{modularity,prunning,pylearn} that discover good combination of hyper-parameters do not employ neuro-evolution, therefore the discovery requirements in terms of time and resources are less expensive, but have other drawbacks such as limited search space, considerable loss in accuracy, limited usability, and need of retraining from scratch for a considerable amount of time after each network alteration.

Other steps towards incremental training are presented in \cite{net2net,networkmorphism}, where the goal is to transfer knowledge from a small network towards a significantly larger network under some architectural constraints. These approaches have a twofold benefit. First, they explore the design space of current state-of-the-art networks in order to find better performers. Second, they avoid information loss when the network structure has to be slightly modified. 
Although they improve state-of-the-art results for the ImageNet classification task, it would be interesting to compare in terms of time/accuracy of the networks obtained through transferred learning with the same networks trained from scratch.

\section{Methodology} \label{sec:methodology}

Let us consider a generic CNN $\mathcal{N}$ composed of $n$~layers. The incremental training method proposed in this work begins by partitioning the original network $\mathcal{N}$ into $K$~sub-networks $\mathcal{S}_k$, with $k=1,...,K$ and $K \leq n$. In our current implementation, the partitioning is done a priori and does not change throughout the training process, although this will be subject of future studies. A sub-network can be composed of one or several layers, with the only constraint that at least one of the layers must contain trainable parameters; in other words, pooling and dropout cannot constitute a sub-network by themselves. The classifier block at the end of the network, which might include fully-connected layers or global pooling layers, is not considered in the $K$~partitions.

\begin{figure*}[!tb]
\centering
\renewcommand{\thesubfigure}{Step 1}
\subfloat[]
{\includegraphics[width=.24\linewidth]{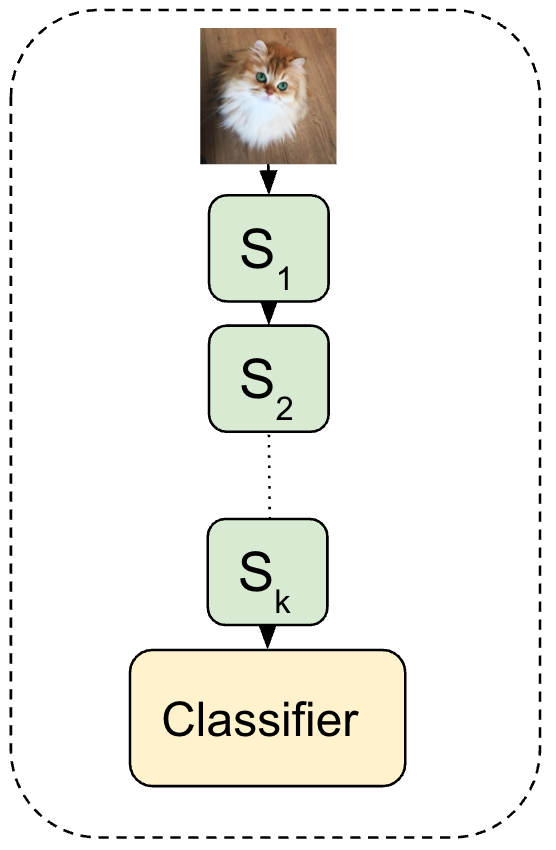}}
\hfil
\renewcommand{\thesubfigure}{Step 2}
\subfloat[]
{\includegraphics[width=.24\linewidth]{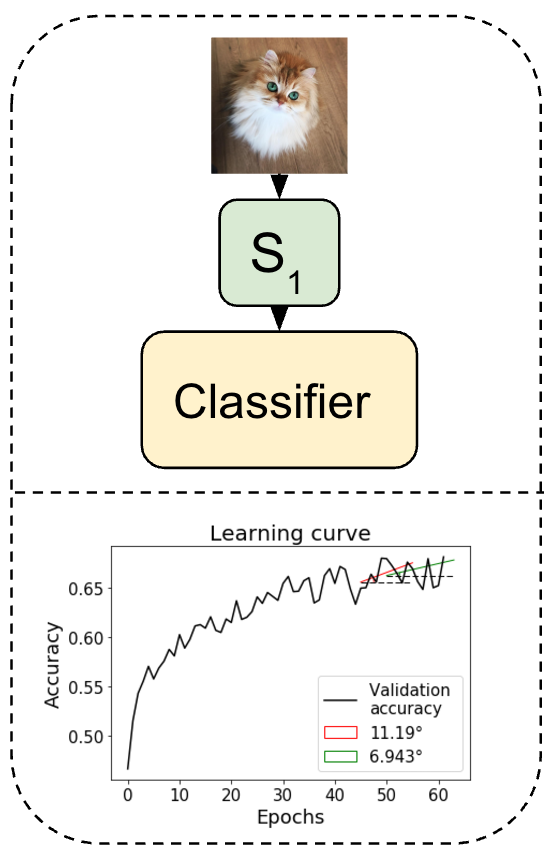}}
\hfill
\renewcommand{\thesubfigure}{Step 3}
\subfloat[]
{\includegraphics[width=.24\linewidth]{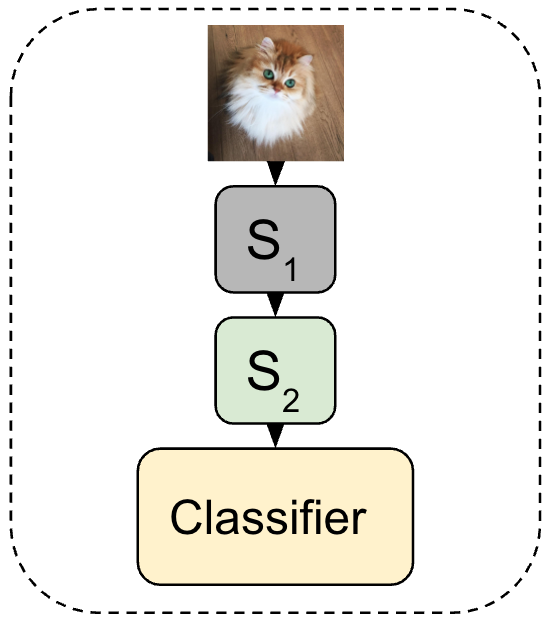}}
\hfill
\renewcommand{\thesubfigure}{Step 4}
\subfloat[]
{\includegraphics[width=.24\linewidth]{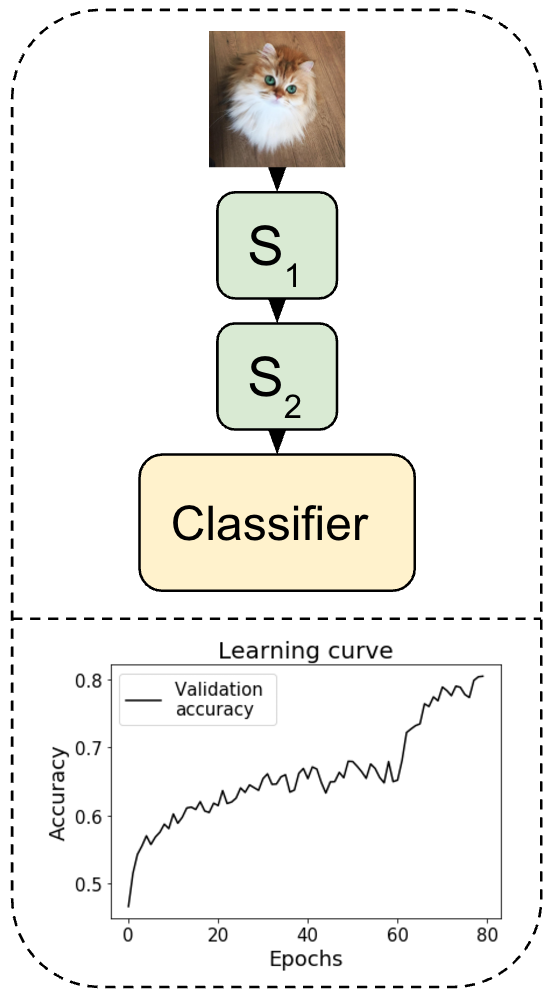}}
\caption{A schematic description of incremental training for a CNN partitioned into $K$ sub-networks. Step~1 depicts the original network that is already split into $K$ sub-networks followed by the classifier layer. The incremental training process starts with the sub-network $\mathcal{S}_{1}$ as shown in Step~2. The learning curve is monitored and when the termination criterion described in Section~\ref{sec:methodology} is met, the training is stopped. $\mathcal{S}_{2}$ is inserted in the current network in Step~3 and is trained only for a few epochs based on the freezed weights of $\mathcal{S}_{1}$. The obtained network $\mathcal{S}_{1}+\mathcal{S}_{2}$ in Step~4 follows the same process as Step~2 and Step~3. The process finishes when either all the sub-networks are incorporated or when a custom criteria is met.}
\vspace*{-5ex}
\label{fig:approach}
\end{figure*}

\begin{figure}[!tb]
\centering
\includegraphics[width=2.7in,height=2in,keepaspectratio]{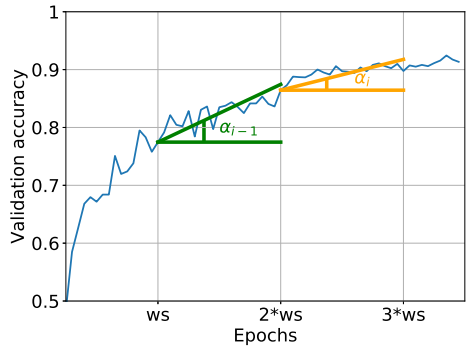}
\caption{Visualization of the termination criteria. Every \textit{window size} ($ws$) epochs we compute the angle between the linear approximation of the last $ws$ accuracy points and the $x$-axis. This angle shows the improvement of the validation accuracy. The training is stopped when $\alpha_{i} \leqslant \gamma \alpha_{i-1} $, where $\gamma$ is a predefined threshold and $\alpha_{i}$ is the angle characterizing the accuracy for the $i$-th window.}
\label{fig:termination_criteria}
\vspace*{-5ex}
\end{figure}

Following the scheme illustrated in Fig.~\ref{fig:approach}, the training process starts with sub-network $\mathcal{S}_1$ attached to the classifier block. To determine when is the optimal time to add the second sub-network $\mathcal{S}_2$ between $\mathcal{S}_1$ and the classifier, we compute every \textit{window size} (ws) epochs the improvement in the validation accuracy. When the improvement observed is below a threshold we stop the training and we increase the network depth by adding the next sub-network. This process, which is illustrated in Fig.~\ref{fig:termination_criteria}, repeats until all $K$~sub-networks are incorporated or until another custom criteria is met.

Each time a new sub-network~$\mathcal{S}_{k+1}$ is inserted in the current architecture, its weights need to be initialized. This step is delicate, as our experiments  show empirically that a non-optimal initialization (e.g., random) might prevent the network from reaching the same accuracy as the original one, or might slow down significantly the incremental training process (Section~\ref{sec:experiments}). To overcome this issue, we propose a more efficient initialization technique, that we refer to as \emph{look-ahead}. Given a current network constituted by $k$~sub-networks, our look-ahead method consists of training $\mathcal{S}_{k+1}$ for a few epochs based on the input generated by the already trained uppermost part of the network. This strategy provides a more informed starting point for the new sub-network, since it looks ahead towards more complex features that can be obtained based on the previously learned ones. We remark that the weights of the original sub-network $\mathcal{S}_{1}+\mathcal{S}_{2}+...+\mathcal{S}_{k}$ are freezed during the look-ahead process that initializes $\mathcal{S}_{k+1}$; this implies that no time is spent in back-propagating in that region of the network. Moreover, the depth of the look-ahead tends to be comparably smaller than the depth of the final network, therefore the training of the look-ahead is not considered expensive.

\section{Experiments} \label{sec:experiments}

In this section, we train several state-of-the-art CNNs to compare the performance of our incremental training method, with the one of the classical algorithm. All runs involve single-precision arithmetic and are performed on IBM\footnote{\fontsize{8}{6}\selectfont IBM, the IBM logo, ibm.com, OpenPOWER are trademarks or registered trademarks of International Business Machines Corporation in the United States, other countries, or both. Other product and service names might be trademarks of IBM or other companies.} POWER8 Minsky compute nodes, equipped with four Nvidia P100 GPUs.

In all experiments we use the CIFAR-10 dataset. CIFAR-10 is a collection of 60,000~RGB images of $32\mathrm{x}32\mathrm{x}3$ pixels each, evenly distributed across 10 mutually exclusive classes. On top of it, we also use the simple data augmentation described in~\cite{lee2015deeply} and summarized in Table~\ref{table:hyper2}, Panel~A. Concerning the baseline networks, we chose ResNet~\cite{resnet} and VGGNet~\cite{vgg}: the former is tailored for CIFAR-10, while the latter is sized for more complex datasets.

\begin{table}[!tb]
\caption{Configuration details adopted during regular and incremental training.}
\label{table:hyper2}
\parbox{.45\linewidth}{
\centering
\begin{tabular}{l@{\,\,\,\,\,\,}l}
\multicolumn{2}{l}{\textbf{Panel A:} Data pre-processing.} \\
\toprule
\multicolumn{1}{l}{Parameter} & \multicolumn{1}{l}{Value} \\
\midrule
Train/Validation split       & 10\%                       \\
Width shift                  & 4 px                   \\
Height shift                 & 4 px                   \\
Horizontal flip              & Yes \\           
  \bottomrule
  &\\
&\\
& \\
\end{tabular}
}
\parbox{.45\linewidth}{
\centering
\begin{tabular}{l@{\,\,\,\,\,\,}l}
\multicolumn{2}{l}{\textbf{Panel B:} Hyper-parameters.} \\
\toprule
\multicolumn{1}{l}{Parameter} & \multicolumn{1}{l}{Value}                                           \\ 
\midrule
Optimizer                    & RMSProp \cite{rmsprop}                                                    \\
Momentum                     & 0.9                                                                  \\
Learning rate                & $10^{-4}$ (no decay)                              \\
\multirow{2}{*}{Batch size}  & 128 (ResNet-X) \\
& 32 (VGG-16) \\
Weight decay                 & $10^{-4}$                                                             \\
Weight initialization        & He \cite{he}     \\    \bottomrule
\end{tabular}
}
\vspace*{-2ex}
\end{table}

The aim of our experiments is to demonstrate the advantages of the incremental training method compared to the regular one. Therefore, in our runs we do not employ all the advanced tricks (e.g.,  learning rate schedule) needed to reach the highest state-of-the-art accuracy. The values of the most relevant hyper-parameters are summarized in Table~\ref{table:hyper2}, Panel~B. The partitioning of the networks for the incremental training is mainly based on simplicity. We group layers with the same number of filters (in the case of convolutions) and the same number of output nodes (in the case of dense layers). 
We tested different types of network partitioning observing no major impact in the overall performance.

\begin{figure}[!tb]
\centering
\captionsetup[subfigure]{labelformat=empty}
\includegraphics[width=2.3in,height=2in,keepaspectratio]{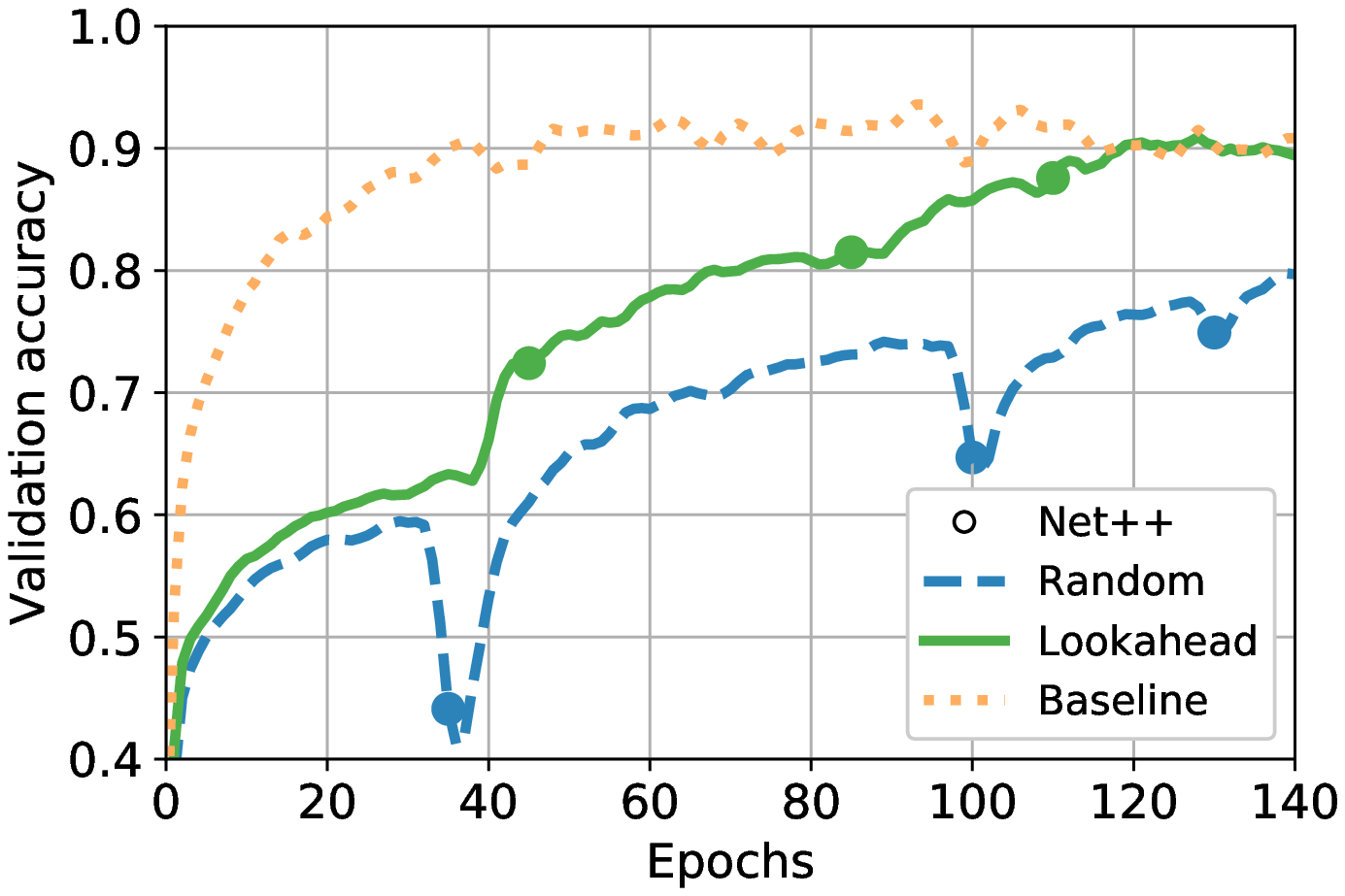}
\includegraphics[width=2.3in,height=2in,keepaspectratio]{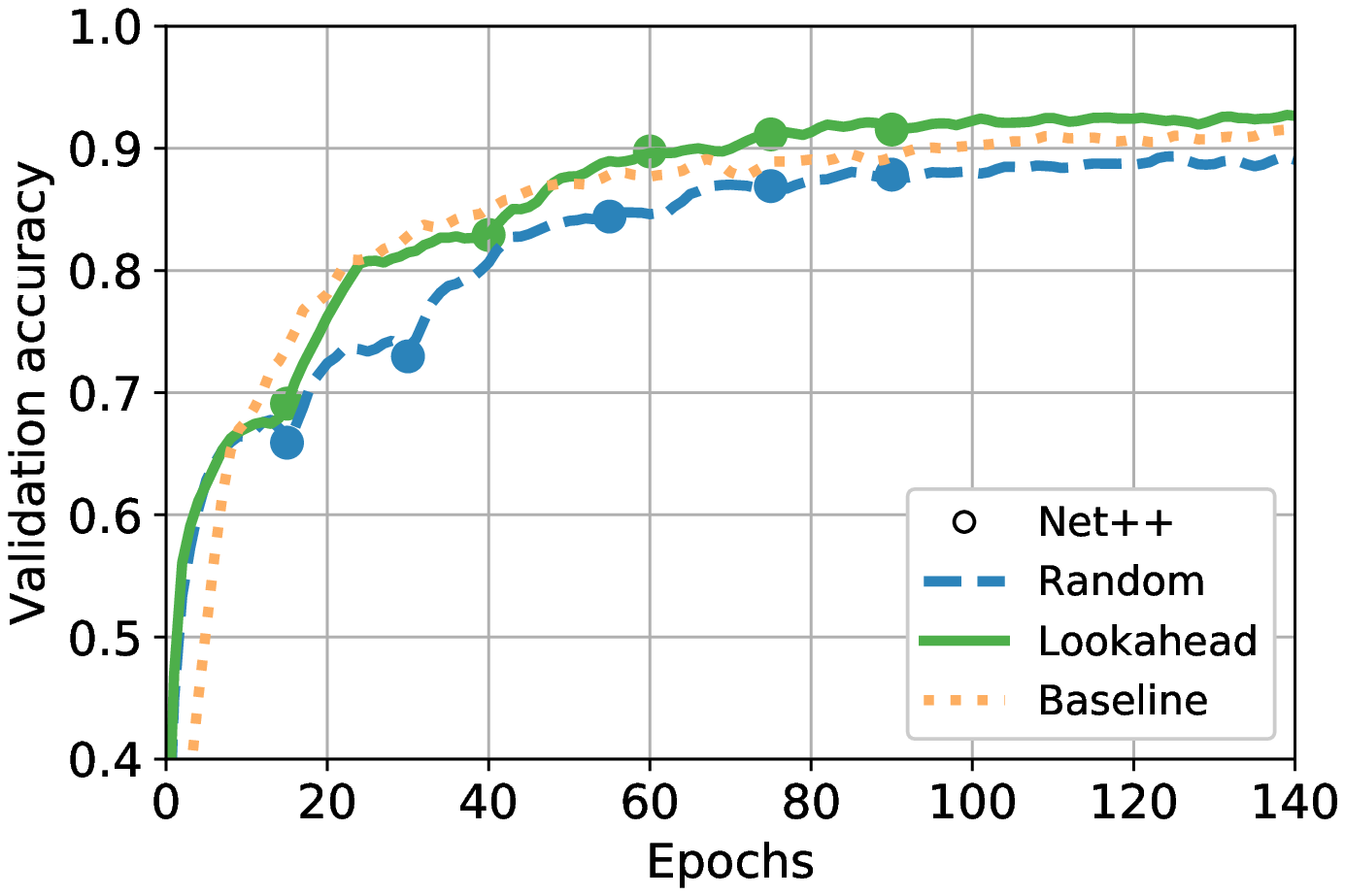}
\includegraphics[width=2.3in,height=2in,keepaspectratio]{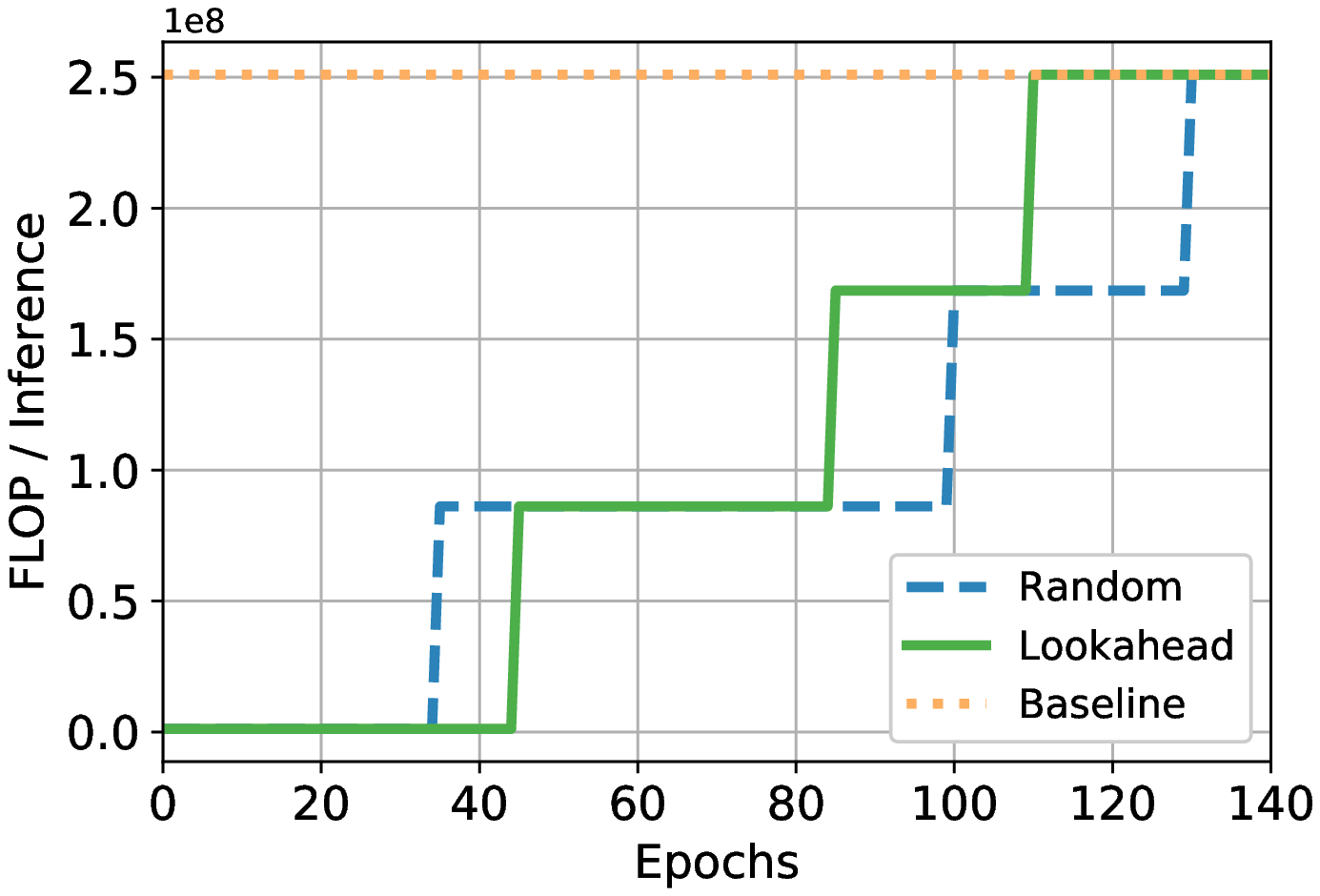}
\includegraphics[width=2.3in,height=2in,keepaspectratio]{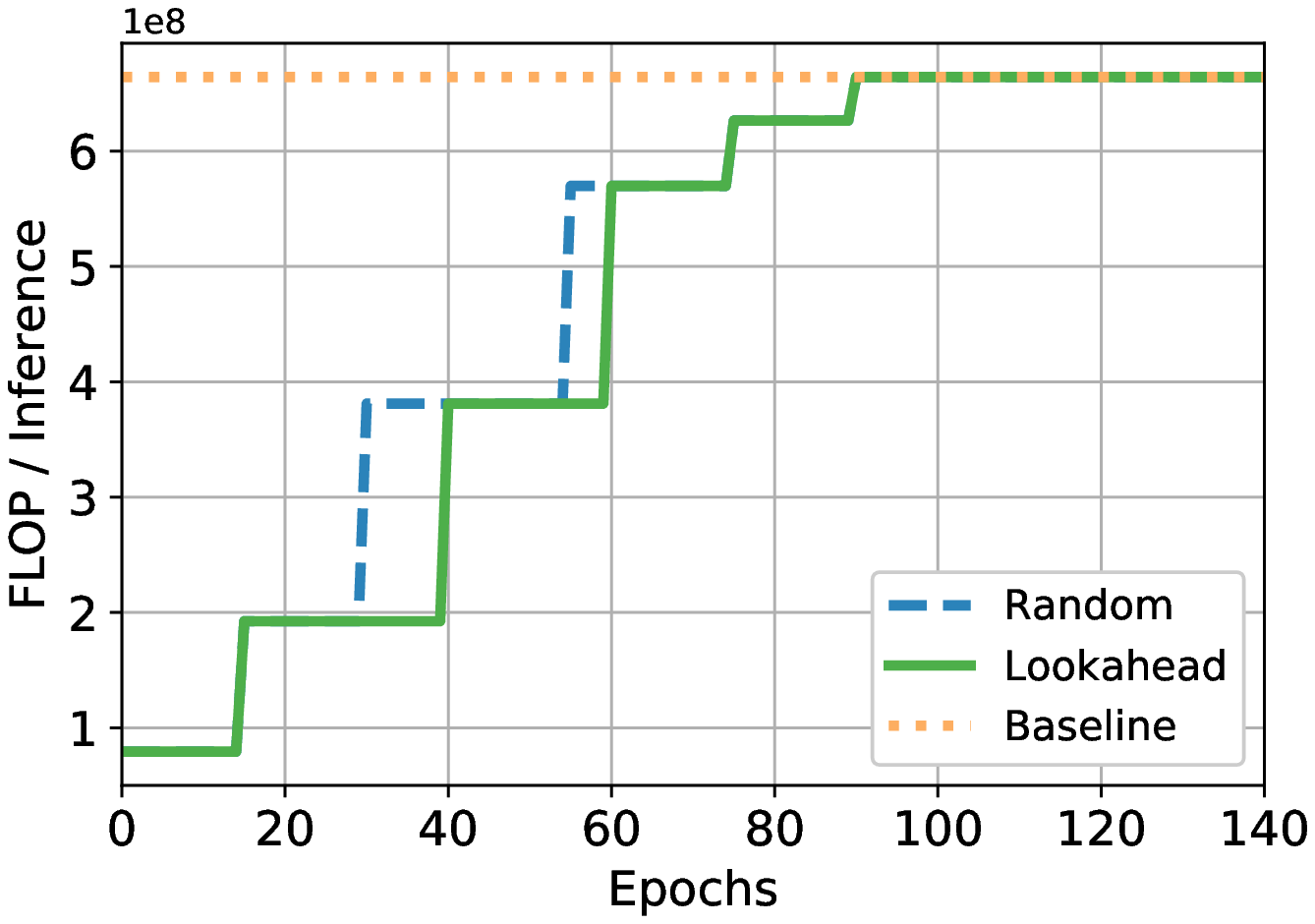}
\subfloat[ResNet-56]{\includegraphics[width=2.3in,height=2in,keepaspectratio]{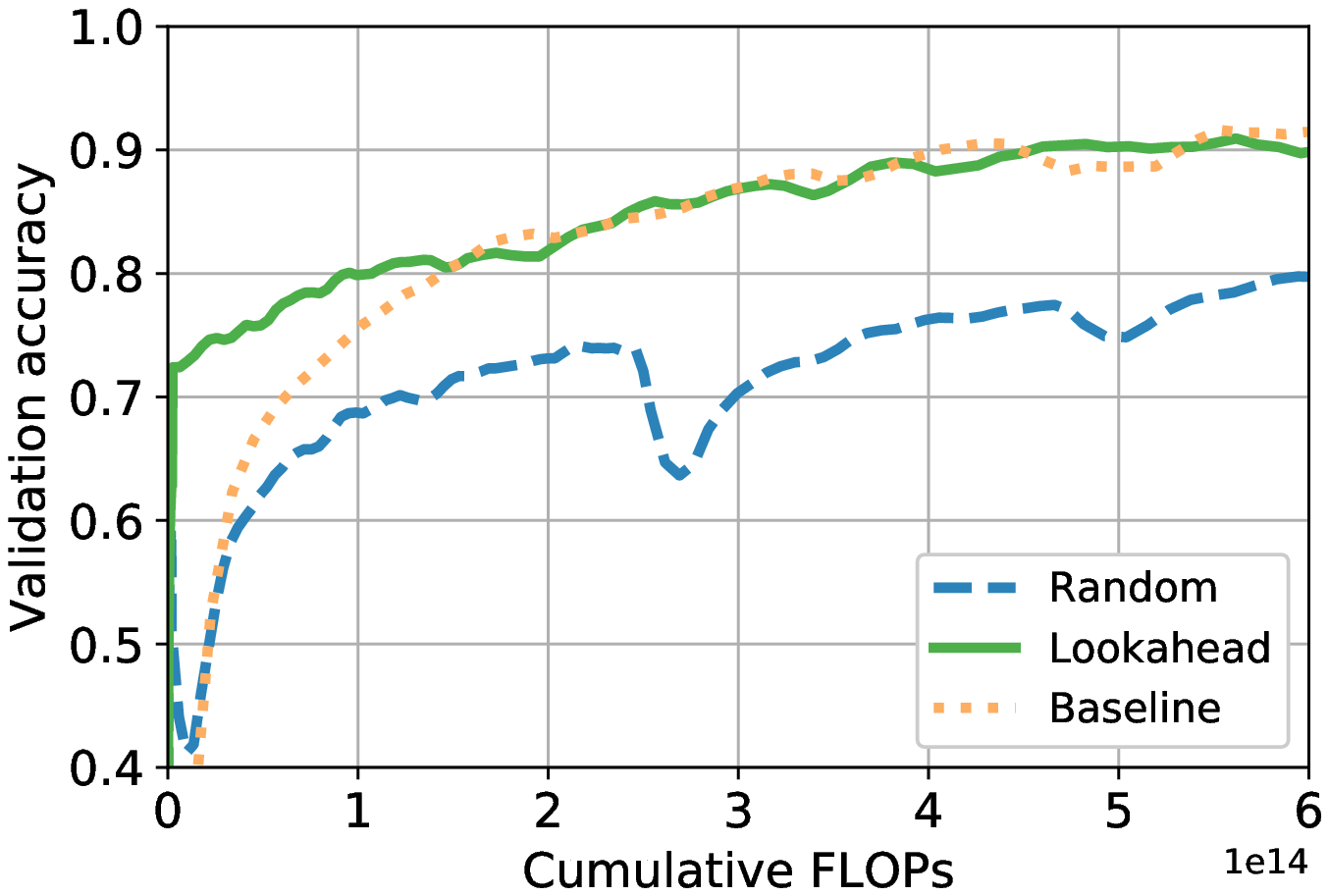}}
\subfloat[VGGNet]{\includegraphics[width=2.3in,height=2in,keepaspectratio]{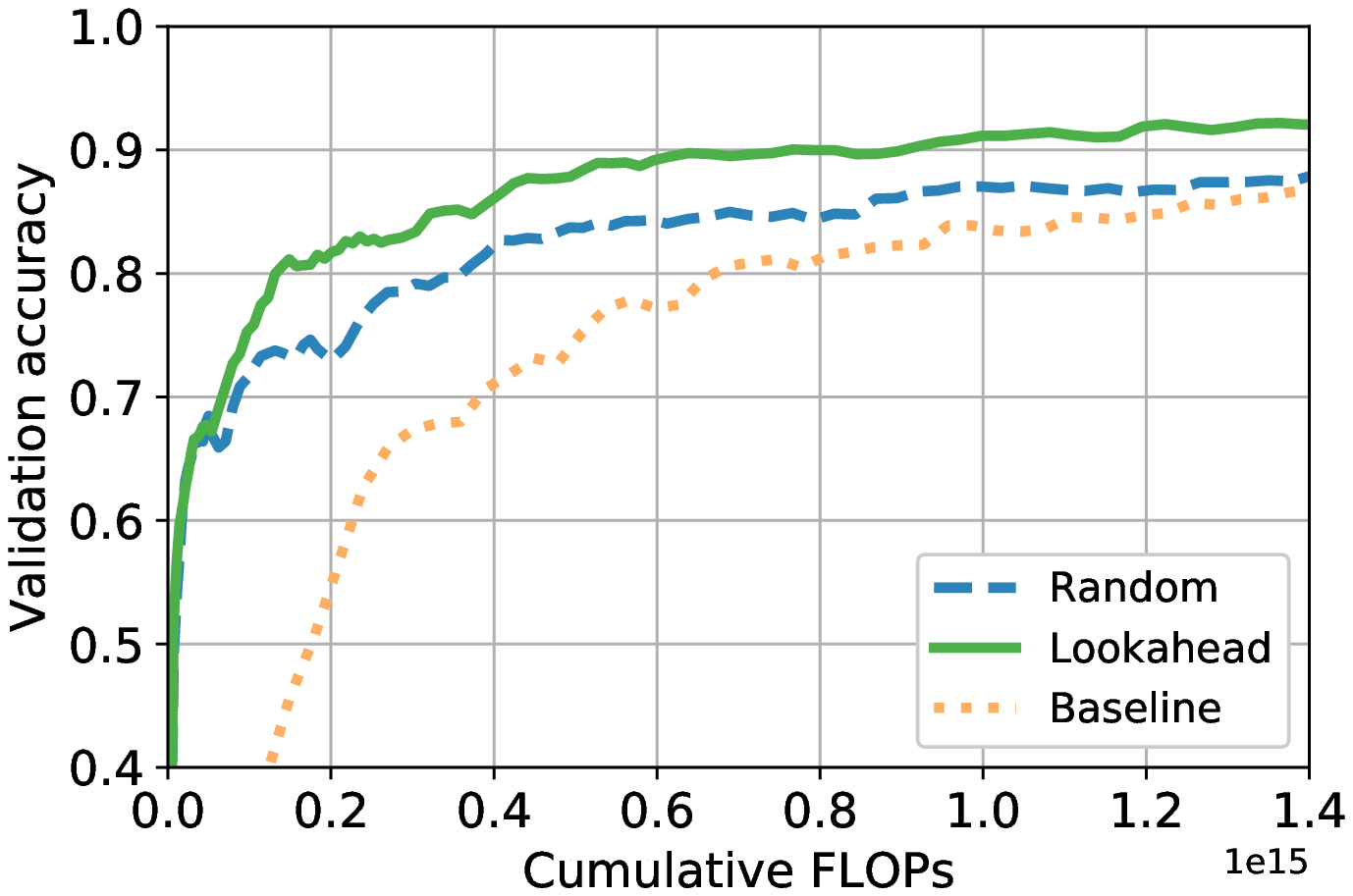}}

\caption{Comparison in terms of accuracy (first row), inference cost (second row), and cumulative FLOPs (third row) between the training of the original network from scratch and its incremental counterpart on the CIFAR-10 dataset.}
\label{fig:vgg_epochs}
\vspace*{-5ex}
\end{figure}

In Fig.~\ref{fig:vgg_epochs} we show a comparison between the regular and incremental training methods: for the latter, we plot results using both random and look-ahead initialization. 
The validation accuracy (first row) shows that the look-ahead outperforms the random initialization of the sub-networks. This behavior is better illustrated in the case of ResNet-56 network, in contrast with the behavior of VGGNet, as the over-parametrized network mitigates the outperformance. 
On the second row we estimate the cost per inference of each sub-network in terms of FLOPs. The plots highlight the benefit of our incremental training, that performs a much lower amount of calculations for the majority of the training time.
On the third and last row we observe that, while the regular version of the VGGNet reached 90\%~accuracy using 1.4~PFLOP, the incremental version needed 0.8~PFLOP and only 40\%~of the total number of parameters. In both networks the look-ahead training converged faster.

\begin{figure}[h!]
  \centering
  \begin{minipage}[t]{0.47\textwidth}
    \includegraphics[width=1.1\textwidth,height=2.4in,keepaspectratio]{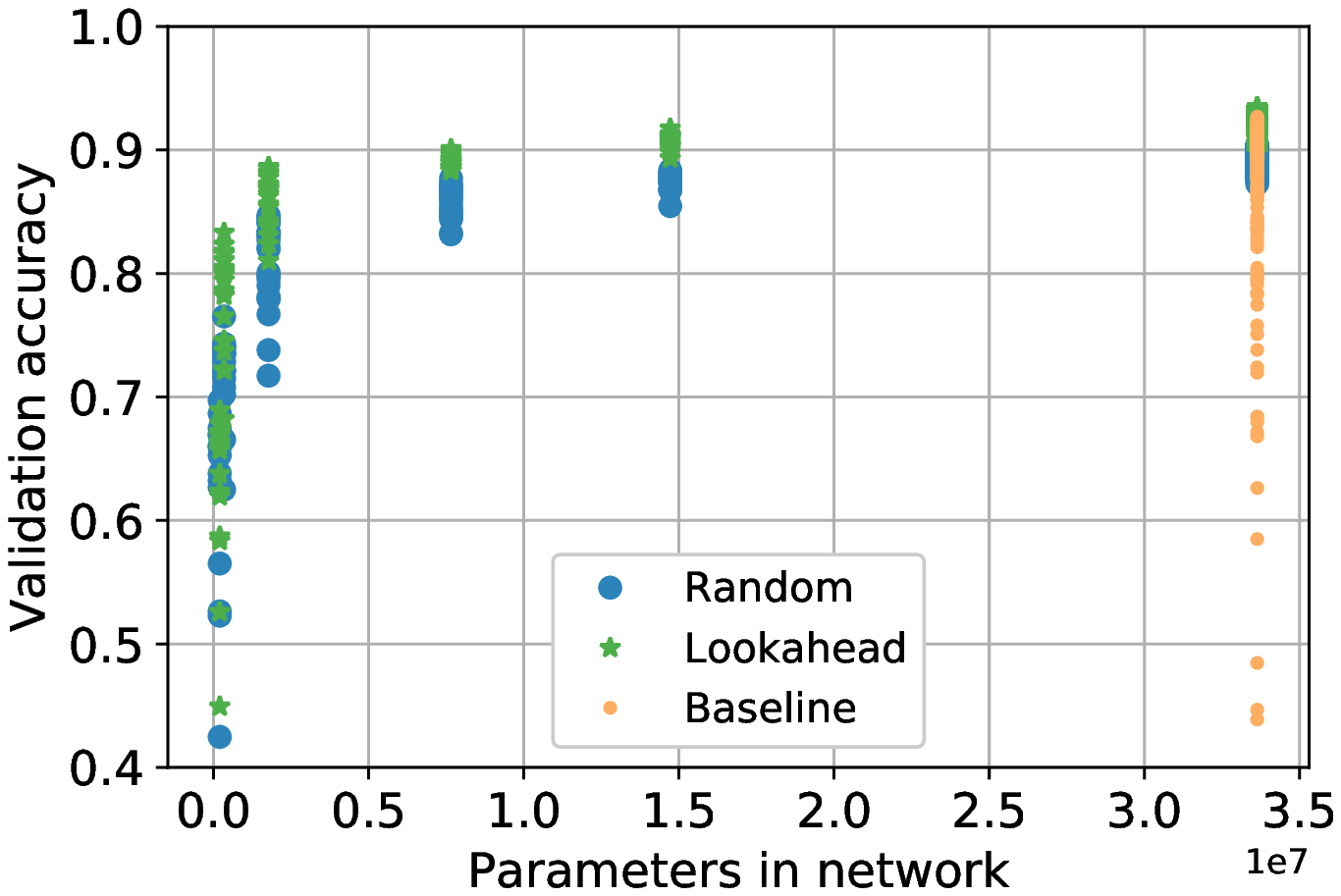}
    \caption{Validation accuracy as a function of the number of parameters in the network. The final accuracy obtained by the VGGNet (which consists of 35M parameters) is reached with only 15M parameters with the incremental training.
    }
    \label{fig:vgg_params}
  \end{minipage}
  \hfill
  \begin{minipage}[t]{0.47\textwidth} 
    \includegraphics[width=1.1\textwidth,height=2.4in,keepaspectratio]{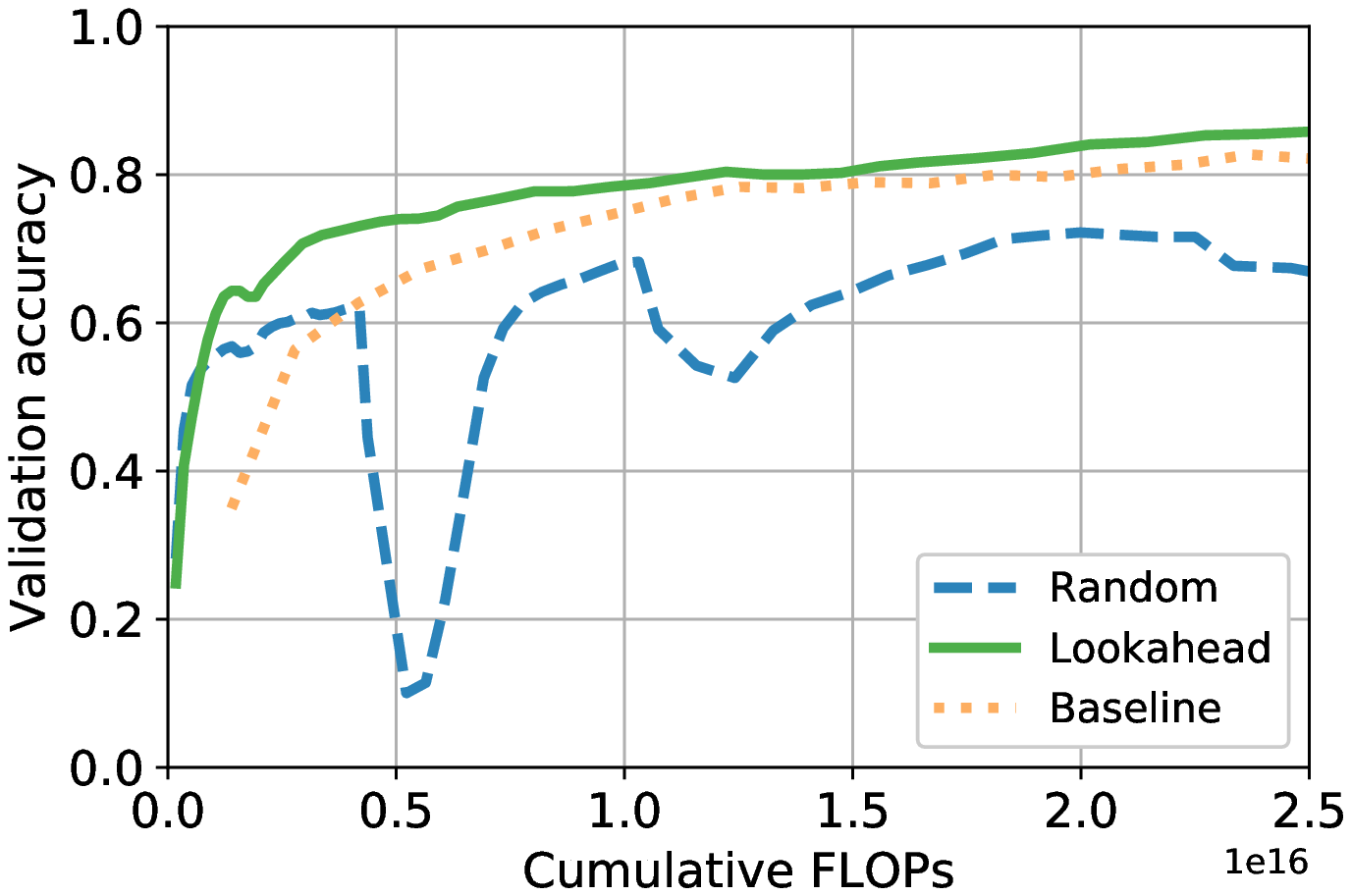}
    \caption{Training of VGGNet on CIFAR-10 images resized to $224\mathrm{x}224\mathrm{x}3$ pixels. For each sub-network we used the maximum possible batch size that fitted the GPU. This allowed for a larger learning rate for the first sub-networks, that led to a faster convergence than the baseline.}
    \label{fig:vgg_params2}
  \end{minipage}
  \vspace*{-9ex}
\end{figure}

On top of these advantages, the incremental training allows to trace the importance of each sub-network during the training. For instance, in the case of VGGNet, the last two sub-networks bring a marginal improvement (less than 1\%~accuracy), while increasing by~16\% the inference cost and by~80\% the number of parameters. In Fig.~\ref{fig:vgg_params} we show that only 5\% of the VGGNet paramaters are enough to reach 85\% accuracy, while with 42\% we arrive to the baseline accuracy. Therefore, the incremental training can be of great benefit for applications in which the last 1-2\% of accuracy are not crucial and a slightly lower performance is within accepted limits. Indeed, it can dynamically determine to stop the network expansion early, saving a lot of computational resources without compromising accuracy.

\begin{table}[!tb]
\centering
\caption{Classification error on CIFAR-10 test set.}
\label{my-label}
\begin{tabular}{l@{\,\,\,\,\,\,}lll}
\toprule
\multirow{2}{*}{Network}       & \multicolumn{3}{c}{Test error(\%)} \\ \cline{2-4} 
                               & \multicolumn{1}{l@{\,\,\,\,\,\,}}{Baseline} & \multicolumn{1}{l@{\,\,\,\,\,\,}}{Look-ahead} & \multicolumn{1}{l@{\,\,\,\,\,\,}}{Random} \\  

\midrule
\multicolumn{1}{l}{VGGNet}   & \multicolumn{1}{c}{90.06}         & \multicolumn{1}{c}{90.50}          & \multicolumn{1}{c}{88.91}       \\ 
\multicolumn{1}{l}{ResNet-20} & \multicolumn{1}{c}{87.81}         & \multicolumn{1}{c}{87.52}          & \multicolumn{1}{c}{79.20}       \\ 
\multicolumn{1}{l}{ResNet-32} & \multicolumn{1}{c}{88.93}         & \multicolumn{1}{c}{88.90}          & \multicolumn{1}{c}{79.89}       \\ 
\multicolumn{1}{l}{ResNet-44} & \multicolumn{1}{c}{89.40}         & \multicolumn{1}{c}{89.32}          & \multicolumn{1}{c}{80.20}       \\ 
\multicolumn{1}{l}{ResNet-56} & \multicolumn{1}{c}{89.51}         & \multicolumn{1}{c}{89.56}          & \multicolumn{1}{c}{80.96}       \\ 
\bottomrule
\end{tabular}
\renewcommand{\arraystretch}{5}
\vspace*{-3ex}
\end{table}

\section{Conclusion} \label{sec:conclusion}

We proposed an incremental training method for CNNs. We demonstrated that our method reaches same or slightly better accuracy than regular training methods on state-of-the-art networks. This was achieved by using a more informed initialization of the network extension, which we call the look-ahead. 
Clear benefits of the incremental training are the faster convergence, the intuitive understanding of the importance of each sub-network in the overall performance, and the smooth synergy between training and optimal network depth discovery. In future works, we plan to generalize the present approach to other types of networks, such as Recurrent Neural Networks and Long Short Term Memory.

\FloatBarrier

\end{document}